\documentclass[letterpaper]{article} 
\usepackage[draft]{aaai2026}  
\usepackage{times}  
\usepackage{helvet}  
\usepackage{courier}  
\usepackage[hyphens]{url}  
\usepackage{graphicx} 
\usepackage{natbib}  
\usepackage{caption} 
\usepackage{amsmath, amssymb, amsthm} 
\usepackage{algorithm}
\usepackage{algorithmic}
\usepackage{booktabs}
\definecolor{CustomGreen}{HTML}{2ca02c}
\definecolor{CustomOrange}{HTML}{fe7f10}
\definecolor{CustomRed}{HTML}{d52628}
\definecolor{CustomBlue}{HTML}{1f77b4}
\usepackage{newfloat}
\usepackage{listings}
\DeclareCaptionStyle{ruled}{labelfont=normalfont,labelsep=colon,strut=off} 
\floatstyle{ruled}
\newfloat{listing}{tb}{lst}{}
\floatname{listing}{Listing}
\title{Towards Inference-time Scaling for Continuous Space Reasoning}
\author {
    Minghan Wang\textsuperscript{\rm 1},
    Thuy-Trang Vu\textsuperscript{\rm 1},
    Ehsan Shareghi\textsuperscript{\rm 1},
    Gholamreza Haffari\textsuperscript{\rm 1}
}
\affiliations {
    \textsuperscript{\rm 1}Department of Data Science \& AI, Monash University\\
    firstname.lastname@monash.edu
}

\begin{document}

\maketitle

\begin{abstract}
Inference-time scaling through multiple sample generation in combination with Process- or Outcome-Reward Model (PRM or ORM) re-ranking has proven effective for text-based reasoning in large language models. This paper investigates whether such established techniques can be successfully adapted to reasoning in the continuous space, using COCONUT~\cite{hao2024traininglargelanguagemodels} continuous space reasoning LM as the backbone. We demonstrate the feasibility of generating diverse reasoning paths through dropout-based sampling. Our Pass@N analysis on the generated samples reveals the potential that could enable a significant gain in performance akin to observed gain in the discrete space. However, we highlight unique challenges faced for materializing this gain in the continuous thought space. In particular, working recipes for data generation and training PRM and ORM models in the discrete space unlocks only marginal improvements in the continuous space. Through probing various aspects including geometric properties and trajectory dynamics we identify the underlying reasons that prevent effective discrimination between correct and incorrect reasoning (essential for the functioning of PRM and ORM). Our findings reveal that current limitations stem from the absence of key inductive biases in continuous thought representations. We argue that the training frameworks for continuous reasoning LMs require not only to optimize for accuracy but also to explicitly incorporate inductive biases that could be utilized during inference-time for discrimination of correct and incorrect thoughts.\footnote{Our code and data will be publicly available.}
\end{abstract}




\section{Introduction}
\begin{figure}[t]
    \centering
    \includegraphics[width=1\columnwidth]{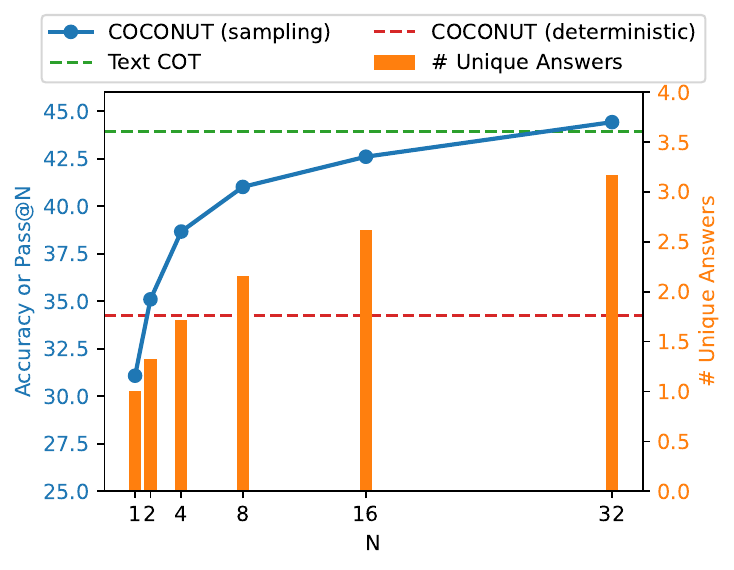}
    \caption{Performance comparison of COCONUT sampling vs. baselines on GSM8k. The \textcolor{CustomBlue}{blue} line shows Pass@N performance with dropout-based sampling, demonstrating substantial potential for inference-time scaling compared to deterministic COCONUT (\textcolor{CustomRed}{red} dashed line) or text CoT (\textcolor{CustomGreen}{green} dashed line). \textcolor{CustomOrange}{Orange} bars indicate the average number of unique answers generated at each sample size, showing logarithmic growth that suggests efficient reranking is feasible.}
    \vspace{-1em}
    \label{fig:accuracy_vs_n}
\end{figure}
Recent advances in reasoning large language models have expanded along training and inference-time dimensions. While training paradigms continue evolving~\citep{openai2024openaio1card,deepseekai2025deepseekr1incentivizingreasoningcapability}, inference-time scaling has stabilized around generating multiple samples per input and employing Process/Outcome Reward Models (PRM/ORM) for re-ranking candidates~\citep{uesato2022solvingmathwordproblems,lightman2023letsverifystepstep,wang2024mathshepherdverifyreinforcellms,wang2023selfconsistencyimproveschainthought} and delivered substantial accuracy improvements across reasoning tasks.

Continuous reasoning~\citep{hao2024traininglargelanguagemodels,dehghani2019universaltransformers,mohtashami2024cotformerchainofthoughtdrivenarchitecture,cheng2024compressedchainthoughtefficient} has emerged as an alternative paradigm that trades interpretability for computational efficiency. Unlike Chain-of-Thought methods generating explicit textual steps~\citep{wei2023chainofthoughtpromptingelicitsreasoning}, continuous reasoning models like COCONUT~\citep{hao2024traininglargelanguagemodels} operate directly in latent space, enabling faster inference while maintaining multi-step reasoning capabilities at the cost of transparency.
This work investigates whether established inference-time scaling techniques from discrete reasoning can be effectively transferred to continuous reasoning models. Specifically, we explore enabling inference-time scaling through multiple sample generation in continuous space and developing reliable reward models to rank continuous thoughts. Using COCONUT as our experimental backbone, we demonstrate that dropout-based sampling can generate diverse reasoning paths, with Pass@N analysis revealing substantial potential to improve overall reasoning accuracy (as shown in Figure~\ref{fig:accuracy_vs_n}) in mathematical reasoning. This indicates a significant room for improvement through effective ranking mechanisms.

Our attempts to train PRM/ORM models adapting established automatic annotation protocols from discrete reasoning, i.e. MATH-Shepherd~\citep{wang2024mathshepherdverifyreinforcellms} yield limited success, falling short of gains observed in discrete (i.e., standard) settings. To understand this limitation, we conduct a comprehensive analysis of the continuous thoughts. Our investigation reveals that continuous thoughts exhibit geometric homogeneity preventing effective discrimination between \emph{correct} and \emph{incorrect} reasoning. Through systematic evaluation, we identify that current limitations stem from the absence of key inductive biases in continuous thought representations during training, where supervision focuses solely on final answer accuracy without promoting structural differentiation in the latent space.

\section{Related Work}

\subsection{Inference-Time Scaling in Discrete Space}
Inference-time scaling aims to enhance the reasoning capability of LLMs at test time by generating and selecting among multiple reasoning paths, without modifying model parameters. Recent works~\citep{lightman2023letsverifystepstep, li202512surveyreasoning} have outlined this as a key direction for enabling more reliable and deliberative reasoning in LLMs.

One foundational strategy is Majority voting (a.k.a. self-consistency)~\citep{wang2023selfconsistencyimproveschainthought}, which improves reasoning by sampling multiple Chain-of-Thoughts and selecting the most consistent answer by picking the most frequent answer. A natural extension of self-consistency is best-of-N sampling with reranking, where multiple reasoning paths are sampled at test time and the final answer is chosen based on scoring criteria such as likelihood, confidence, or verifier output. These methods include structured prompting approaches such as Least-to-Most~\citep{zhou2023leasttomostpromptingenablescomplex}, and iterative refinement or search-based techniques like Self-Refine~\citep{madaan2023selfrefineiterativerefinementselffeedback} and Tree-of-Thoughts~\citep{yao2023treethoughtsdeliberateproblem}, illustrating the range of reranking mechanisms across different granularity levels.

Another line of inference-time methods employs verifier models to evaluate and rank sampled reasoning paths based on their correctness or plausibility, often at the step level rather than just final answers. To assess the correctness of these intermediate steps, Process Reward Models have been introduced. PRMs assign a score to reasoning steps, reflecting their correctness likelihood. One of the key challenges in training such models is the lack of step-level human annotations (with the exception of PRM800k dataset by \citet{lightman2023letsverifystepstep}), which is often circumvented via automatic annotation methods. Among existing methods, Math-Shepherd~\citep{wang2024mathshepherdverifyreinforcellms} is commonly used as a de-facto approach for automatically generating step-level supervision from existing CoT outputs. There are other approaches that propose alternative means for more reliable supervision~\cite{zhang2025lessonsdevelopingprocessreward}.

\subsection{Continuous Space Reasoning}
Continuous CoT performs multi-step inference directly in a model’s latent space, bypassing explicit textual steps~\citep{sui2025stopoverthinkingsurveyefficient}. Instead of generating intermediate tokens, the model refines hidden representations across steps, aiming for more efficient and expressive reasoning~\citep{zhu2025surveylatentreasoning}.
Several methods implement this idea. 
CODI~\citep{shen2025codicompressingchainofthoughtcontinuous} improves latent stability by aligning hidden states between student and teacher models using self-distillation. CCOT~\citep{cheng2024compressedchainthoughtefficient} compresses multi-step reasoning into a variable-length sequence of latent embeddings, optionally decodable for interpretability. Token Assorted~\citep{su2025tokenassortedmixinglatent} introduces a hybrid approach that combines discrete latent tokens generated by VQ-VAE with text tokens. CoT2~\citep{gozeten2025continuouschainthoughtenables} demonstrates that models are able to track multiple traces in parallel when reasoning in continuous space and proposes strategies to explicitly encourage such parallel reasoning.

Among these methods, COCONUT~\citep{hao2024traininglargelanguagemodels} represents a prominent work in continuous reasoning that enables models to perform reasoning directly in hidden state space rather than through explicit textual steps. During training, the model employs multi-stage curriculum learning where, at training stage $k$, the first $k$ reasoning steps are replaced with $k \times c$ continuous thought vectors (where $c$ controls the number of latent thoughts per reasoning step), progressively transitioning from language-based to latent-based reasoning. During inference, the reasoning process can be formalized as follows: given a problem prompt $X$, a special \texttt{<bot>} token initiates continuous reasoning mode. The model generates a sequence of continuous thought vectors $\{\mathbf{s}_1, ... , \mathbf{s}_T\}$ through autoregressive computation:
\begin{equation}
    \mathbf{s}_i = f_{\theta}(X, \mathbf{s}_{<i})
\end{equation}
where $f_{\theta}$ represents the model's forward pass. Crucially, these hidden states are directly fed as input for subsequent reasoning steps without decoding to text space. After $T$ predetermined reasoning steps, an \texttt{
<eot>} token terminates the continuous phase, and the model returns to standard text generation to produce the final answer.

In this work, we adopt COCONUT as our backbone to systematically investigate whether established inference-time scaling techniques in the text space can be effectively transferred to continuous reasoning models. To the best of our knowledge, our work is the first in exploring inference-time scaling for continuous space reasoning.

\section{Generating Continuous Thought Samples}


Inference-time scaling in text-based LLMs relies on sampling multiple reasoning trajectories from token probability distributions at each generation step, ensuring trajectory diversity while preserving reasoning coherence for techniques like self-consistency voting.

COCONUT presents a fundamental challenge: its reasoning process operates deterministically within continuous space. While sampling can be applied during final answer generation, all samples originate from identical reasoning trajectories, failing to achieve the path diversity necessary for effective inference-time scaling.

To address this limitation, we introduce a simple yet effective approach that injects controlled stochasticity into the continuous reasoning process. We selectively enable dropout during the iterative hidden state generation phase while disabling it during the text generation phase for answer production. This design confines randomness to the reasoning process without compromising final answer generation integrity.

\subsection{Preliminary Experiments}
\subsubsection{Experimental Setup}
We strictly follow the original paper~\citep{hao2024traininglargelanguagemodels}  to reproduce experiments on GSM8k~\citep{cobbe2021trainingverifierssolvemath}. Specifically, we use GPT-2~\citep{radford2019language} as the base model and train for 6 epochs in the initial stage and 3 epochs in each of the remaining stages (3 stages total). During inference, we employ $T = 3 \times c$ where $(c=2)$ continuous thought steps, identical to the configuration in the original paper.

For dropout-based sampling, we use the same dropout rate as during training and enable dropout only during continuous reasoning. We evaluate with sample sizes of $\{1,2,4,8,16,32\}$. We establish two baselines: text CoT and the standard deterministic COCONUT reasoning, reporting their accuracy scores. For the sampling evaluation, we report Pass@N metrics where $N$ equals the sample size, i.e. a problem is considered correct if any of the $N$ sampled answers is correct, representing the  upper-bound achievable by the model. We also report the number of unique answers after deduplicating the $N$ candidates.

\subsubsection{Experimental Results}

Figure~\ref{fig:accuracy_vs_n} reveals that deterministic COCONUT reasoning exhibits nearly a 10-point accuracy gap compared to text CoT, consistent with the original COCONUT paper. In our dropout-based sampling evaluation, Pass@1 is the single-sample accuracy and is slightly lower than deterministic COCONUT reasoning, indicating that enabling dropout during inference introduces a modest degradation in reasoning quality. Despite the performance impact at $N=1$, Pass@N rapidly surpasses the deterministic COCONUT baseline as sample size increases, ultimately exceeding the text CoT baseline at $N=32$. Critically, the number of unique answers exhibits logarithmic rather than linear growth with increasing $N$. This observation suggests that effective re-ranking methods could achieve substantial accuracy gains by identifying correct answers from the generated candidate set, while maintaining computational efficiency since re-ranking costs do not scale linearly with sample size if applying additional deduplication.

These preliminary findings motivated us to further explore the possibility of training the reward models to fully exploit the sampled solutions via re-ranking.

\section{Reward Modeling for Continuous Thought}
\begin{table*}[]
\centering
\resizebox{\textwidth}{!}{%
\begin{tabular}{@{}l|ccc|ccc|ccc@{}}
\toprule
\textbf{N} & \textbf{\begin{tabular}[c]{@{}c@{}}\# Unique\\ Answer\end{tabular}} & \textbf{\begin{tabular}[c]{@{}c@{}}\# Correct \\ Answers\end{tabular}} & \textbf{\begin{tabular}[c]{@{}c@{}}\# Major \\ Incorrect Answers\end{tabular}} & \textbf{Pass@N} & \textbf{Confidence} & \textbf{Self-Consistency} & \textbf{PRM-HE} & \textbf{PRM-SE} &  \textbf{ORM} \\ \midrule
\textbf{1} & 1.00 & 0.31 & 0.69 & 31.08 & 31.08 & 31.08 & 31.08 & 31.08 & 31.08 \\
\textbf{2} & 1.33 & 0.62 & 1.14 & 35.10 & 31.08 & 31.01 & \textbf{31.69} & 30.86 & 31.61 \\
\textbf{4} & 1.72 & 1.24 & 2.11 & 38.67 & 30.48 & 31.61 & \textbf{32.45} & 32.37 & 32.15 \\
\textbf{8} & 2.16 & 2.45 & 4.09 & 41.02 & 29.87 & 31.24 & \textbf{33.06} & 32.52 & 31.46 \\
\textbf{16} & 2.62 & 4.95 & 8.02 & 42.61 & 31.39 & 32.15 & \textbf{33.36} & 32.37 & 32.37 \\
\textbf{32} & 3.17 & 9.84 & 15.88 & 44.43 & 30.71 & 32.15 & 32.83 & \textbf{33.28} & 31.39 \\ \bottomrule
\end{tabular}%
}
\caption{BoN Performance comparison of different reranking methods on GSM8k. Pass@N shows theoretical upper bounds, while PRM and ORM achieve modest improvements over baseline methods (Confidence, Self-Consistency) but fall short of the potential indicated by oracle selection. Best results among reranking methods are highlighted in bold.}
\label{tab:PRM_ORM_overall_results}
\end{table*}

Building on our preliminary findings that demonstrate significant potential for inference-time scaling in continuous reasoning, we address the critical challenge of effectively ranking model-generated candidates.
We adopt the well-established \emph{discrete} data annotation framework from MATH-Shepherd~\citep{wang2024mathshepherdverifyreinforcellms} to construct training data for continuous thought process supervision and develop both process reward models (PRM) and outcome reward models (ORM) specifically designed for COCONUT. We examine their effectiveness in identifying and ranking correct reasoning trajectories.
\begin{algorithm}[H]
\caption{MC Annotation for COCONUT Thoughts}
\label{alg:annotation}
\begin{algorithmic}[1]
\REQUIRE Problem prompt $X$, ground truth answer $a^*$, \\
number of trajectories $M$, number of MC completions $N$, reasoning steps $T$
\ENSURE Step-wise rewards $\{y^{SE}_{s_i}, y^{HE}_{s_i}\}_{i=1}^{|\mathcal{T}_{unique}|}$ for all reasoning trajectories

\STATE \textbf{Generate reasoning trajectories:}
\STATE $\mathcal{T} \leftarrow \{\tau_1, \ldots, \tau_M\}$ where $\tau_m = \{\mathbf{s}_1^{(m)}, \ldots, \mathbf{s}_T^{(m)}\}$
\STATE Each $\mathbf{s}_i^{(m)} \in \mathbb{R}^D$ is a continuous thought vector from COCONUT

\STATE \textbf{Deduplicate trajectories:}
\STATE $\mathcal{T}_{unique} \leftarrow \text{Deduplicate}(\mathcal{T})$ based on final answers

\STATE \textbf{Annotate each reasoning step:}
\FOR{each trajectory $\tau \in \mathcal{T}_{unique}$}
    \FOR{$i = 1$ to $T$}
        \STATE $\mathbf{s}_i \leftarrow \tau[i]$ \COMMENT{Current reasoning step}
        \STATE $\tau_{1:i} \leftarrow \{\mathbf{s}_1, \mathbf{s}_2, \ldots, \mathbf{s}_i\}$ \COMMENT{Partial trajectory}
        
        \STATE \textbf{Monte Carlo estimation:} \\
        \COMMENT{Generate $N$ completions from $\tau_{1:i}$}
        \STATE $\{a_j\}_{j=1}^N \leftarrow \text{Complete}(\tau_{1:i}, N)$
        \STATE $y^{HE}_{s_i} \leftarrow \mathbf{1}[\exists j: a_j = a^*]$ \COMMENT{Hard estimation}
        \STATE $y^{SE}_{s_i} \leftarrow \frac{\sum_{j=1}^N \mathbf{1}[a_j = a^*]}{N}$ \COMMENT{Soft estimation}
    \ENDFOR
\ENDFOR

\RETURN Step-wise reward labels for PRM training
\end{algorithmic}
\end{algorithm}
\subsection{Data Curation}

MATH-Shepherd introduces an automated data annotation methodology using Monte Carlo (MC) estimation that circumvents expensive human annotation by estimating reasoning step success probabilities through multiple sampling. We adapt this framework to annotate continuous thought vectors for reward model training. 

The annotation process can be found in Algo~\ref{alg:annotation}. Given a problem prompt $X$, we employ the trained COCONUT model to generate multiple reasoning trajectories (line 2), treating each continuous thought vector as an individual reasoning step. In our configuration, where COCONUT generates $T = 6$ latent vectors per trajectory, we annotate all 6 vectors independently.

For each reasoning step $s_i$ in a trajectory, we generate $N$ candidate completions from that step (line 11) and evaluate their final answers $\{a_j\}_{j=1}^N$ against the ground truth answer $a^*$. The hard estimation (line 13) of the step-wise rewards is computed as:
\begin{align}
    y_{s_i}^{HE} = \begin{cases} 
    1 & \exists a_j \in \{a_1, ..., a_N\}, a_j = a^* \\
    0 & \text{Otherwise}
    \end{cases}
\end{align}
and the soft estimation (line 14) is computed as:
\begin{equation}
    y_{s_i}^{SE} = \frac{\sum_{j=1}^{N} \mathbb{I}(a_j = a^*)}{N}
\end{equation}

The hard estimation assigns a positive label if any completion from step $s_i$ leads to the correct answer, indicating that the step preserves the potential for correct reasoning. For trajectory-level evaluation, we compute outcome rewards as: $r^{OUT} = \mathbb{I}[\text{final answer is correct}]$.

\subsection{Modeling}

A fundamental distinction between continuous and text-based reasoning lies in data transferability across architectures. In text-based reward modeling, data generation and reward models can use different architectures since both operate in a shared textual space. However, continuous thought representations are model-specific, in which only the originating model can interpret its own latent reasoning steps. This restricts us to using COCONUT itself as the backbone for both PRM and ORM.

We augment the COCONUT backbone with task-specific prediction heads: classification heads for PRM hard labels and ORM outcome labels, and a regression head for PRM soft labels. Each head consists of two linear layers with ReLU activation in between. A sigmoid function is applied on all heads to map output values $\in [0,1]$. For PRM training, we employ joint loss combining cross-entropy for hard estimation and mean squared error for soft estimation:
\begin{equation}
    \mathcal{L}_{PRM} = \mathcal{L}_{CE}(y_{s_i}^{HE}, \hat{y}_{s_i}^{HE}) + \mathcal{L}_{MSE}(y_{s_i}^{SE}, \hat{y}_{s_i}^{SE})
\end{equation}
For ORM training, we use cross-entropy loss only: 
\begin{equation}
    \mathcal{L}_{ORM} = \mathcal{L}_{CE}(r^{OUT}, \hat{r}^{OUT})
\end{equation}

\subsection{Experiments}

\subsubsection{Experimental Setup}
For RM training data construction, we utilize the GSM8K training set, generating M=5 reasoning paths per problem and deduplicating based on final answers (resulting in 1.32 samples per problem on average). We annotate every trajectory step regardless of answer correctness (maintaining consistency with the MATH-Shepherd open-sourced dataset~\footnote{https://huggingface.co/datasets/peiyi9979/Math-Shepherd} to ensure complete trajectory coverage) and generate N=10 completions for MC estimation during step-wise annotation.

To ensure training stability, we balance positive and negative samples maintaining a 1:1 ratio: PRM uses step-wise hard estimation labels while ORM uses outcome labels. The resulting training sets contain 238k samples for PRM and 324k samples for ORM. Both models are trained for 10 epochs with peak learning rate 1e-4 (500 warmup steps), batch size 128, on a single A100 GPU.

For evaluation, we employ Best-of-N (BoN) methodology~\citep{stiennon2022learningsummarizehumanfeedback} with $N \in \{1, 2, 4, 8, 16, 32\}$ on the GSM8k test set, retaining all candidates without deduplication to assess genuine ranking effectiveness. We compare RM-based verification against confidence-based reranking using answer probability scores and self-consistency via majority voting~\citep{wang2023selfconsistencyimproveschainthought}. Beyond accuracy metrics, we analyze the average number of unique answers, correct answers among N candidates, and majority incorrect answers to provide comprehensive evaluation insights.
\begin{table}[]
\centering
\resizebox{\columnwidth}{!}{%
\begin{tabular}{@{}l|cccc|ccccc@{}}
\toprule
 & \multicolumn{4}{c|}{\textbf{Hard Estimation}} & \multicolumn{4}{c}{\textbf{Soft Estimation}} \\ \midrule
\textbf{N} & \textbf{last} & \textbf{min} & \textbf{max} & \textbf{mean} & \textbf{last} & \textbf{min} & \textbf{max} & \textbf{mean} \\ \midrule
\textbf{1} & 31.08 & 31.08 & 31.08 & 31.08 & 31.08 & 31.08 & 31.08 & 31.08 \\
\textbf{2} & 31.69 & 31.16 & 31.16 & 31.39 & 30.86 & 30.55 & 31.08 & 30.86 \\
\textbf{4} & 32.45 & 32.30 & 31.92 & 32.07 & 32.37 & 32.22 & 31.69 & 32.52 \\
\textbf{8} & 33.06 & 32.98 & 31.69 & 33.06 & 32.52 & 31.46 & 32.22 & 33.06 \\
\textbf{16} & 33.36 & 33.13 & 32.37 & 33.13 & 32.37 & 31.61 & 31.77 & 32.52 \\
\textbf{32} & 32.83 & 32.52 & 31.84 & 32.90 & 33.28 & 32.60 & 32.75 & 33.06 \\ \bottomrule
\end{tabular}%
}
\caption{Performance of BoN using different aggregating strategies reranked with the PRM.}
\label{tab:PRM_ablation_aggragating}
\vspace{-1em}
\end{table}

\subsubsection{Experimental Results}

As shown in Table~\ref{tab:PRM_ORM_overall_results}, Pass@N upper bounds demonstrate substantial potential for inference-time scaling, consistent with our preliminary study (Figure~\ref{fig:accuracy_vs_n}). However, model-intrinsic approaches prove inadequate: confidence-based reranking provides no improvement, indicating poor calibration in COCONUT, while self-consistency shows marginal gains but remains limited. The underlying issue becomes apparent when examining answer distributions—across all N values, correct answers consistently fall below dominant incorrect answers, revealing COCONUT's inability to achieve effective scaling through intrinsic capabilities alone.

Both PRM and ORM outperform model-intrinsic methods, with PRM-HE achieving the most consistent improvements, reaching 33.36\% accuracy at N=16 compared to the 31.08\% baseline. While this confirms reward models' potential utility, the improvement magnitude is limited, less than 2.3 points despite a theoretical upper bound of 42.61\%. These gains pale compared to text-based RM verification, where similar methodologies yield substantially larger improvements.

Investigation of different score aggregation strategies (min, max, mean, last-step) following~\citet{zhang2025lessonsdevelopingprocessreward} shows negligible variation across methods (Table~\ref{tab:PRM_ablation_aggragating}), suggesting that core limitations transcend simple scoring mechanisms and necessitate deeper analysis of reward model capabilities.

\section{Analysis}

The substantial gap between reward model reranking performance and theoretical upper bounds motivates a detailed investigation into the root causes of reward model ineffectiveness. We hypothesize two potential explanations for this limitation. First, the reward models may suffer from fundamental discriminative capacity constraints, failing to reliably distinguish between correct and incorrect reasoning paths or final answers. To examine this possibility, we conduct a systematic evaluation of reward model classification performance. Second, the  distribution of COCONUT's continuous reasoning trajectories may exhibit weak correlation with answer correctness, preventing reward models from extracting meaningful discriminative features. To investigate this hypothesis, we analyze COCONUT's continuous reasoning from three complementary perspectives: geometric properties of thought representations, trajectory dynamics, and behavioral responses under perturbation.

\subsection{Curation of Evaluation Dataset}

For subsequent experiments, we construct a dedicated test set for systematic evaluation based on the GSM8k test set, employing the same annotation methodology used for training data construction (as detailed in Algo~\ref{alg:annotation}). To ensure high-quality annotations, we increase the number of generated samples per problem to $M=10$ (retaining an average of 2.29 answers per problem after deduplication) and expand the MC estimation candidate size to $N=20$ for enhanced label reliability. The resulting evaluation dataset contains 3,014 samples with 18.48\% correct answers and 28.21\% correct reasoning steps. We deliberately preserve the natural distribution without rebalancing to accurately reflect reward model performance across the full spectrum of correct and incorrect reasoning patterns.

\begin{table}[]
\centering
\resizebox{\columnwidth}{!}{%
\begin{tabular}{@{}lccccc@{}}
\toprule
\textbf{Model} & \textbf{Accuracy} & \textbf{Precision} & \textbf{Recall} & \textbf{F1-Score} & \textbf{Specificity} \\ \midrule
\textbf{PRM} & 62.98 & 41.60 & 77.28 & 54.09 & 57.36 \\
\textbf{ORM} & 73.72 & 39.11 & 75.76 & 51.59 & 73.26 \\ \bottomrule
\end{tabular}%
}
\caption{Classification performance of PRM and ORM.}
\label{tab:prm_orm_classification}
\vspace{-1em}
\end{table}

\subsection{Classification Performance of RMs}

We evaluate the classification performance of both PRM (hard estimation prediction) and ORM using our constructed test set, employing a threshold of 0.5 to distinguish between correct and incorrect steps or solutions for both models.

\begin{figure}
    \centering
    \includegraphics[width=1.0\linewidth]{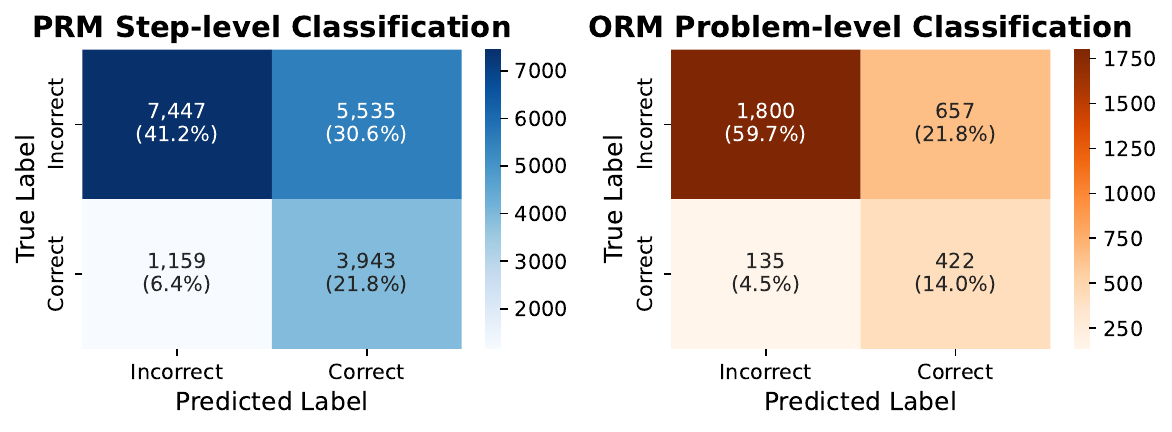}
    \caption{Confusion matrix of PRM and ORM.}
    \label{fig:prm_orm_cm}
    \vspace{-1.0em}
\end{figure}

As shown in Table~\ref{tab:prm_orm_classification}, both reward models exhibit poor classification performance with notably low F1-scores (PRM: 54.09\%, ORM: 51.59\%), indicating fundamental discriminative limitations. The confusion matrices in Figure~\ref{fig:prm_orm_cm} reveal that PRM suffers from substantial false positive errors: 5,535 false positives (30.6\%) against only 3,943 true positives (21.8\%), resulting in poor precision (41.60\%) despite reasonable recall (77.28\%). This pattern demonstrates that PRM frequently misclassifies incorrect reasoning steps as correct, failing to reliably distinguish between valid and invalid continuous thoughts. ORM achieves better overall accuracy (73.72\%) and specificity (73.26\%) but maintains similarly low precision (39.11\%), indicating persistent misclassification issues across both models.

These classification deficiencies directly explain the limited reranking effectiveness observed in our experiments. The consistently low F1-scores and poor precision fundamentally constrain the models' ability to identify and prioritize correct reasoning paths, suggesting deeper structural challenges inherent to continuous space reward modeling that cannot be addressed through simple methodological adjustments.

\subsection{Geometric Properties of Thought Representation}
In this analysis, we treat thought vectors as independent representations and examine their geometric properties using two key metrics to understand their spatial distribution characteristics.

We first consider isotropy, which measures the uniformity of variance across all dimensions, revealing how thought vectors distribute and utilize dimensions in high-dimensional space. We employ IsoScore$\star$ proposed by~\citet{rudman2024stableanisotropicregularization} due to its superior properties, including mean agnosticism, rotation invariance, and global stability. IsoScore$\star$ operates by computing the covariance matrix eigenvalues $\Lambda=\{\lambda_1, ..., \lambda_d\}$ of input thought vectors $\mathbf{S}$, and normalizing them as $\hat{\Lambda} = \sqrt{d} \cdot \Lambda/||\Lambda||_2$, where $d$ represents the dimensionality of thought vectors. The final score is calculated as:
\begin{align}
    \delta(\Lambda) &= \frac{||\hat{\Lambda} - \mathbf{1}||}{\sqrt{2(d-\sqrt{d})}}, \quad \mathbf{1} = (1,...,1)^T \in \mathbb{R}^d \nonumber\\
    \phi(\Lambda) &= \frac{(d - \delta(\Lambda)^2(d-\sqrt{d}))^2}{d^2} \nonumber\\
    \text{IsoScore}\star &= \frac{d \cdot \phi(\Lambda) - 1}{d-1}
\end{align}
A value of 1 indicates perfect isotropy, while 0 indicates maximal anisotropy.

We also examine sparsity using the Hoyer~\citep{hurley2009comparingmeasuressparsity} metric to assess dimensional activation patterns:
\begin{equation}
    Hoyer(s_i) = \frac{\sqrt{d} - ||s_i||_1/||s_i||_2}{\sqrt{d-1}}
\end{equation}
Higher Hoyer values indicate sparser activations where fewer dimensions carry primary information, corresponding to more focused reasoning representations, while lower values suggest distributed activation patterns.

We evaluate both metrics across three sample groups: all thoughts in the test set (noted as Entire Set), thoughts correctly predicted by PRM (where PRM prediction matches ground truth, noted as PRM+), and thoughts incorrectly predicted by PRM (where PRM prediction differs from ground truth, noted as PRM-). Within each group, we separately analyze thoughts labeled as ``correct" and ``incorrect."

\begin{figure}
    \centering
    \includegraphics[width=1.0\linewidth]{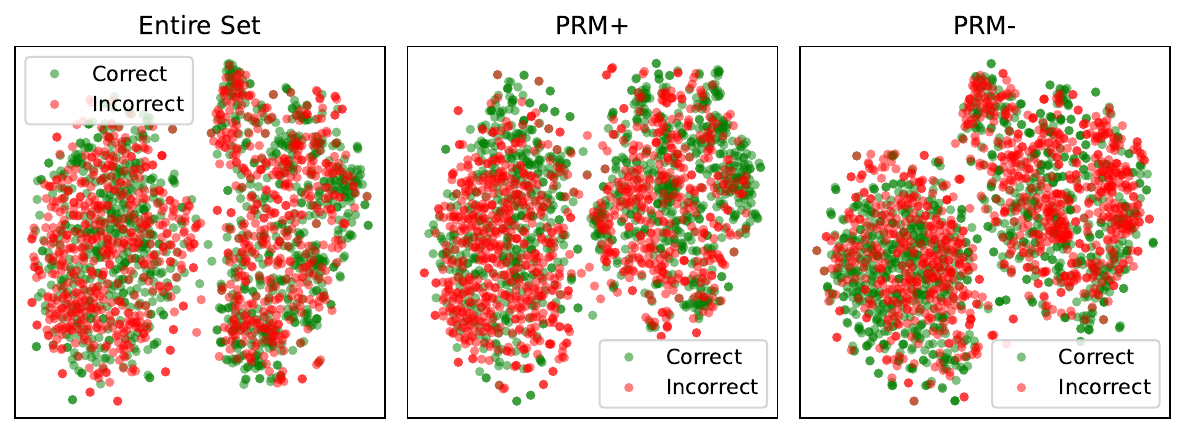}
    \caption{t-SNE of the sampled latent thoughts labeled with correct/incorrect for the entire set, PRM+ and PRM- groups.}
    \label{fig:tsne}
\end{figure}

Table~\ref{tab:geo_properties} reveals several critical findings: (1) Thought representations exhibit low isotropy and relative sparsity, confirming that reasoning operates within limited dimensional subspaces. (2) Most importantly, across all measurement groups, the differences between correct and incorrect thoughts are negligible, indicating that geometric properties (present in COCONUT thought vectors) alone cannot distinguish reasoning step correctness. This fundamental lack of geometric separability explains why PRM fails to effectively discriminate through these features. The t-SNE visualization in Figure~\ref{fig:tsne} further confirms this observation, showing that correct and incorrect thoughts are completely intermixed in the representation space~\footnote{The two distinct clusters are due to 2 vectors per reasoning step ($c=2$), where the model differentiates between vector positions within each step. This clustering is unrelated to reasoning correctness.}.

\begin{table}[]
\centering
\resizebox{\columnwidth}{!}{%
\begin{tabular}{l|cc|cc}
\hline
\textbf{Metrics} & \multicolumn{2}{c|}{\textbf{IsoScore$\star$}} & \multicolumn{2}{c}{\textbf{Hoyer}} \\ \hline
\textbf{Groups} & \textbf{Correct} & \textbf{Incorrect} & \textbf{Correct} & \textbf{Incorrect} \\ \hline
\textbf{Entire set} & 0.0134 & 0.013 & 0.21 ± 0.01 & 0.22 ± 0.01 \\
\textbf{PRM+} & 0.0137 & 0.0131 & 0.21 ± 0.01 & 0.22 ± 0.01 \\
\textbf{PRM-} & 0.0126 & 0.0132 & 0.22 ± 0.01 & 0.21 ± 0.01 \\ \hline
\end{tabular}%
}
\caption{Geometric properties with Hoyer and IsoScore$\star$. Results are grouped with the entire set, PRM+ (samples correctly predicted by the PRM), PRM- (samples incorrectly predicted by the PRM). Values are computed within subsets of thought vectors annotated with their correctness based on the hard estimation.}
\label{tab:geo_properties}
\vspace{-1em}
\end{table}
\begin{table}[ht]
\centering
\resizebox{\columnwidth}{!}{%
\begin{tabular}{@{}lcccc@{}}
\toprule
\textbf{Metric} & \textbf{Correct} & \textbf{Incorrect} & \textbf{$p$-value} & \textbf{Cohen's $d$} \\
\midrule
\multicolumn{5}{c}{\textbf{Entire Set}} \\
\midrule
\textbf{Compactness}      & 19.81 $\pm$ 2.53 & 19.39 $\pm$ 2.48 & \textbf{0.023}* & $0.17$ \\
\textbf{Curvature}        & 9.32 $\pm$ 0.52 & 9.38 $\pm$ 0.53 & 0.161 & $-0.10$ \\
\textbf{Local Smoothness} & 0.43 $\pm$ 0.11 & 0.44 $\pm$ 0.11 & 0.074 & $-0.13$ \\
\textbf{Straightness}     & 0.22 $\pm$ 0.04 & 0.21 $\pm$ 0.04 & 0.637 & $0.04$ \\
\midrule
\multicolumn{5}{c}{\textbf{ORM+}} \\
\midrule
\textbf{Compactness}      & 19.91 $\pm$ 2.26 & 19.56 $\pm$ 2.42 & 0.295 & $0.15$ \\
\textbf{Straightness}     & 0.22 $\pm$ 0.04 & 0.21 $\pm$ 0.05 & 0.553 & $0.09$ \\
\textbf{Local Smoothness} & 0.42 $\pm$ 0.10 & 0.45 $\pm$ 0.11 & 0.089 & $-0.24$ \\
\textbf{Curvature}        & 9.30 $\pm$ 0.42 & 9.40 $\pm$ 0.50 & 0.169 & $-0.20$ \\
\midrule
\multicolumn{5}{c}{\textbf{PRM+}} \\
\midrule
\textbf{Compactness}      & 20.72 $\pm$ 1.97 & 18.55 $\pm$ 1.83 & \textbf{0.022}* & $1.14$ \\
\textbf{Straightness}     & 0.22 $\pm$ 0.03 & 0.23 $\pm$ 0.03 & 0.490 & $-0.32$ \\
\textbf{Local Smoothness} & 0.39 $\pm$ 0.09 & 0.48 $\pm$ 0.10 & \textbf{0.049}* & $-0.97$ \\
\textbf{Curvature}        & 8.87 $\pm$ 0.59 & 8.90 $\pm$ 0.52 & 0.884 & $-0.07$ \\
\bottomrule
\end{tabular}%
}
\caption{Comparison of geometric metrics between correct and incorrect reasoning chains on the entire set, ORM+ subset (ORM correctly predicted samples), and PRM+ subset (PRM correctly predicted samples). Bold $p$-values indicate statistical significance ($p < 0.05$).}
\label{tab:trajectory_dynamics}
\vspace{-1em}
\end{table}

\subsection{Trajectory Dynamics}

In this analysis, we examine COCONUT's reasoning process from a trajectory perspective to investigate whether reasoning correctness is reflected in trajectory dynamics. We measure reasoning trajectories across four dimensions: spatial distribution, path shape, local continuity, and global efficiency.

We first examine \textbf{compactness}, which quantifies how tightly trajectory points cluster around their centroid:
\begin{align}
    \overline{\mathbf{s}} = \frac{1}{T} \sum_{i=1}^{T} \mathbf{s}_i \quad
    R_g^2 &= \frac{1}{T} \sum_{i=1}^{T} \|\mathbf{s}_i - \overline{\mathbf{s}}\|_2^2 \nonumber \\
    \text{compactness} &= \sqrt{R_g^2}
\end{align}
where $T$ is the number of reasoning steps, $\mathbf{s}_i$ represents the $i$-th thought vector, and $\overline{\mathbf{s}}$ is the trajectory centroid. Smaller compactness values indicate trajectories that remain within a concentrated region of reasoning space.
Next, we measure \textbf{curvature}~\citep{hosseini2023largelanguagemodelsimplicitly}, capturing the total bending energy along the reasoning path:
\begin{align}
    \mathbf{\Delta}_i = \mathbf{s}_{i+1} - \mathbf{s}_i \quad
    \theta_i &= \arccos\left(\frac{\mathbf{\Delta}_{i-1} \cdot \mathbf{\Delta}_i}{\|\mathbf{\Delta}_{i-1}\|_2 \|\mathbf{\Delta}_i\|_2}\right), i \geq 2 \nonumber \\
    \text{curvature} &= \sum_{i=2}^{T-1} \theta_i
\end{align}
where $\mathbf{\Delta}_i$ represents the displacement vector between consecutive thoughts, and $\theta_i$ is the angle between successive displacement vectors.
We assess \textbf{local smoothness} by quantifying coherence between consecutive reasoning steps using cosine similarity:
\begin{align}
    \text{local\_smoothness} &= \frac{1}{T-1} \sum_{i=1}^{T-1} \frac{\mathbf{s}_i \cdot \mathbf{s}_{i+1}}{\|\mathbf{s}_i\|_2 \|\mathbf{s}_{i+1}\|_2}
\end{align}
Finally, we evaluate \textbf{straightness}, measuring global path efficiency by comparing net displacement to total path length:
\begin{align}
    L = \sum_{i=1}^{T-1} \|\mathbf{s}_{i+1} - \mathbf{s}_i\|_2 \quad
    D &= \|\mathbf{s}_{T} - \mathbf{s}_1\|_2 \nonumber \\
    \text{straightness} &= \frac{D}{L} \in [0, 1]
\end{align}
where $L$ is the total path length and $D$ is the net displacement from start to end. Values approaching 1 indicate efficient, direct reasoning paths, while lower values suggest circuitous or exploratory trajectories.

Similar to our geometric analysis, we analyze three sample groups: all trajectories in the test set, trajectories correctly predicted by ORM, and trajectories where PRM correctly predicts all steps. For each group, we calculate metrics for trajectories labeled as ``correct" or ``incorrect" and perform t-tests.

As shown in Table~\ref{tab:trajectory_dynamics}, most metrics show minimal differences between correct and incorrect trajectories. While compactness shows statistical significance in the entire set ($p=0.023$) and PRM+ subset ($p=0.022$), the effect sizes remain small (Cohen's d=0.17 and 1.14 respectively). The most notable finding occurs in the PRM+ subset, where correct trajectories exhibit higher compactness and lower local smoothness, potentially suggesting that correct reasoning paths are more spatially concentrated yet less smooth between consecutive steps. Overall, the results demonstrate that correct and incorrect reasoning trajectories exhibit no substantial differences in trajectory dynamics, further limiting the possibility of RMs to leverage these characteristics for effective discrimination.

\subsection{Analysis under Perturbation}

Given the lack of significant geometric differences between correct and incorrect thought vectors, we investigate whether COCONUT's reasoning paths contain necessary semantic information or merely serve as positional placeholders. We design a perturbation experiment injecting Gaussian noise into latent thoughts at varying intensities (i.e. $\text{ratio}\times\text{noise}+(1-\text{ratio})\times\text{thought}$) while observing Pass@5 (with sampling size $N=5$) performance on GSM8k. Metrics such as the average number of unique/correct answers among the 5 candidates, and the percentage of problems where the majority answer remains unchanged compared to the previous noise level are also computed, serving as a measurement of answer stability.

As shown in Table~\ref{tab:noise_perterbation}, COCONUT demonstrates robustness at low noise ratios (0.0-0.2), with minimal degradation and 76\% of majority answers unchanged. This aligns with our findings of high anisotropy (i.e., low isotropy) and sparsity—noise primarily affects irrelevant dimensions while preserving critical reasoning dimensions.

Performance degradation becomes pronounced at higher noise ratios, but strikingly, even with complete noise corruption (ratio=1.0), Pass@5 remains non-zero at 12.59\%. This suggests COCONUT's reasoning does not exclusively depend on latent thoughts—for many problems, the model can generate correct answers independently of continuous reasoning. This raises questions about continuous thoughts' actual contribution.

\section{Discussion}
Through systematic analysis of reward model performance, geometric properties, and perturbation, we uncover fundamental challenges in reward modeling for continuous thought models. Our findings reveal that continuous thoughts cluster within concentrated high-dimensional space without sufficient semantic differentiation. This prevents effective discrimination between correct and incorrect reasoning trajectories.

We attribute this clustering to the absence of inductive biases or geometric structural constraints during COCONUT's training, where supervision applies only to text tokens while latent thoughts are generated without explicit guidance. This results in homogeneous thought vectors that learn shared characteristics without encoding distinctive semantic properties, confounding reward model training.

Our analysis suggests that introducing targeted inductive biases during training could potentially address these limitations. Promising directions include encouraging higher isotropy in thought representations, promoting trajectory diversity through varying geometric patterns, and incorporating contrastive learning to teach discrimination between correct and incorrect reasoning patterns. By establishing clearer geometric structure in continuous thought space, these approaches could enable effective reward modeling and unlock inference-time scaling potential.

\begin{table}[]
\centering
\resizebox{\columnwidth}{!}{%
\begin{tabular}{lcccc}
\hline
\textbf{\begin{tabular}[c]{@{}l@{}}Noise \\ Ratio\end{tabular}} & \textbf{\begin{tabular}[c]{@{}c@{}}\# Unique \\ Answer\end{tabular}} & \textbf{Pass@5} & \textbf{\begin{tabular}[c]{@{}c@{}}\# Correct\\ Answer\end{tabular}} & \textbf{\begin{tabular}[c]{@{}c@{}}\% Majority Answer \\ Unchange\end{tabular}} \\ \hline
\textbf{0.0} & 1.86 & 39.20 & 1.55 & 100.00 \\
\textbf{0.2} & 1.92 & 38.67 & 1.50 & 76.35 \\
\textbf{0.4} & 2.22 & 34.80 & 1.24 & 67.32 \\
\textbf{0.6} & 2.56 & 20.32 & 0.55 & 44.73 \\
\textbf{0.8} & 2.62 & 15.62 & 0.37 & 49.43 \\
\textbf{1.0} & 2.49 & 12.59 & 0.32 & 53.83 \\ \hline
\end{tabular}%
}
\caption{Noise perturbation results in different noise levels.}
\label{tab:noise_perterbation}
\end{table}
\section{Conclusion}
In this work, we investigate inference-time scaling for continuous reasoning models through multiple sample generation and reward model reranking using COCONUT. While preliminary studies demonstrate promising potential with substantial Pass@N improvements, adapting established text-based reward modeling methodologies yields limited success. Through comprehensive analysis, we identify that continuous thoughts lack geometric separability necessary for effective discrimination, with correct and incorrect reasoning representations exhibiting minimal structural differences. Our findings highlight that unlocking inference-time scaling in continuous reasoning requires training frameworks that promote geometric differentiation in the continuous thought space, establishing both the potential and challenges for future research.

\clearpage
\bibliography{aaai2026}

\begin{thebibliography}{27}
\providecommand{\natexlab}[1]{#1}

\bibitem[{Cheng and Durme(2024)}]{cheng2024compressedchainthoughtefficient}
Cheng, J.; and Durme, B.~V. 2024.
\newblock Compressed Chain of Thought: Efficient Reasoning Through Dense Representations.
\newblock arXiv:2412.13171.

\bibitem[{Cobbe et~al.(2021)Cobbe, Kosaraju, Bavarian, Chen, Jun, Kaiser, Plappert, Tworek, Hilton, Nakano, Hesse, and Schulman}]{cobbe2021trainingverifierssolvemath}
Cobbe, K.; Kosaraju, V.; Bavarian, M.; Chen, M.; Jun, H.; Kaiser, L.; Plappert, M.; Tworek, J.; Hilton, J.; Nakano, R.; Hesse, C.; and Schulman, J. 2021.
\newblock Training Verifiers to Solve Math Word Problems.
\newblock arXiv:2110.14168.

\bibitem[{DeepSeek-AI et~al.(2025)DeepSeek-AI, Guo, Yang, Zhang, and et~al.}]{deepseekai2025deepseekr1incentivizingreasoningcapability}
DeepSeek-AI; Guo, D.; Yang, D.; Zhang, H.; and et~al., J.~S. 2025.
\newblock DeepSeek-R1: Incentivizing Reasoning Capability in LLMs via Reinforcement Learning.
\newblock arXiv:2501.12948.

\bibitem[{Dehghani et~al.(2019)Dehghani, Gouws, Vinyals, Uszkoreit, and Łukasz Kaiser}]{dehghani2019universaltransformers}
Dehghani, M.; Gouws, S.; Vinyals, O.; Uszkoreit, J.; and Łukasz Kaiser. 2019.
\newblock Universal Transformers.
\newblock arXiv:1807.03819.

\bibitem[{Gozeten et~al.(2025)Gozeten, Ildiz, Zhang, Harutyunyan, Rawat, and Oymak}]{gozeten2025continuouschainthoughtenables}
Gozeten, H.~A.; Ildiz, M.~E.; Zhang, X.; Harutyunyan, H.; Rawat, A.~S.; and Oymak, S. 2025.
\newblock Continuous Chain of Thought Enables Parallel Exploration and Reasoning.
\newblock arXiv:2505.23648.

\bibitem[{Hao et~al.(2024)Hao, Sukhbaatar, Su, Li, Hu, Weston, and Tian}]{hao2024traininglargelanguagemodels}
Hao, S.; Sukhbaatar, S.; Su, D.; Li, X.; Hu, Z.; Weston, J.; and Tian, Y. 2024.
\newblock Training Large Language Models to Reason in a Continuous Latent Space.
\newblock arXiv:2412.06769.

\bibitem[{Hosseini and Fedorenko(2023)}]{hosseini2023largelanguagemodelsimplicitly}
Hosseini, E.~A.; and Fedorenko, E. 2023.
\newblock Large language models implicitly learn to straighten neural sentence trajectories to construct a predictive representation of natural language.
\newblock arXiv:2311.04930.

\bibitem[{Hurley and Rickard(2009)}]{hurley2009comparingmeasuressparsity}
Hurley, N.~P.; and Rickard, S.~T. 2009.
\newblock Comparing Measures of Sparsity.
\newblock arXiv:0811.4706.

\bibitem[{Li et~al.(2025)Li, Zhang, Zhang, Zhang, Liu, Yao, Xu, Zheng, Wang, Chen, Zhang, Yin, Dong, Li, Bi, Mei, Fang, Liang, Guo, Song, and Liu}]{li202512surveyreasoning}
Li, Z.-Z.; Zhang, D.; Zhang, M.-L.; Zhang, J.; Liu, Z.; Yao, Y.; Xu, H.; Zheng, J.; Wang, P.-J.; Chen, X.; Zhang, Y.; Yin, F.; Dong, J.; Li, Z.; Bi, B.-L.; Mei, L.-R.; Fang, J.; Liang, X.; Guo, Z.; Song, L.; and Liu, C.-L. 2025.
\newblock From System 1 to System 2: A Survey of Reasoning Large Language Models.
\newblock arXiv:2502.17419.

\bibitem[{Lightman et~al.(2023)Lightman, Kosaraju, Burda, Edwards, Baker, Lee, Leike, Schulman, Sutskever, and Cobbe}]{lightman2023letsverifystepstep}
Lightman, H.; Kosaraju, V.; Burda, Y.; Edwards, H.; Baker, B.; Lee, T.; Leike, J.; Schulman, J.; Sutskever, I.; and Cobbe, K. 2023.
\newblock Let's Verify Step by Step.
\newblock arXiv:2305.20050.

\bibitem[{Madaan et~al.(2023)Madaan, Tandon, Gupta, Hallinan, Gao, Wiegreffe, Alon, Dziri, Prabhumoye, Yang, Gupta, Majumder, Hermann, Welleck, Yazdanbakhsh, and Clark}]{madaan2023selfrefineiterativerefinementselffeedback}
Madaan, A.; Tandon, N.; Gupta, P.; Hallinan, S.; Gao, L.; Wiegreffe, S.; Alon, U.; Dziri, N.; Prabhumoye, S.; Yang, Y.; Gupta, S.; Majumder, B.~P.; Hermann, K.; Welleck, S.; Yazdanbakhsh, A.; and Clark, P. 2023.
\newblock Self-Refine: Iterative Refinement with Self-Feedback.
\newblock arXiv:2303.17651.

\bibitem[{Mohtashami, Pagliardini, and Jaggi(2024)}]{mohtashami2024cotformerchainofthoughtdrivenarchitecture}
Mohtashami, A.; Pagliardini, M.; and Jaggi, M. 2024.
\newblock CoTFormer: A Chain-of-Thought Driven Architecture with Budget-Adaptive Computation Cost at Inference.
\newblock arXiv:2310.10845.

\bibitem[{OpenAI et~al.(2024)OpenAI, :, Jaech, Kalai, Lerer, Richardson, El-Kishky, and et~al.}]{openai2024openaio1card}
OpenAI; :; Jaech, A.; Kalai, A.; Lerer, A.; Richardson, A.; El-Kishky, A.; and et~al., A.~L. 2024.
\newblock OpenAI o1 System Card.
\newblock arXiv:2412.16720.

\bibitem[{Radford et~al.(2019)Radford, Wu, Child, Luan, Amodei, and Sutskever}]{radford2019language}
Radford, A.; Wu, J.; Child, R.; Luan, D.; Amodei, D.; and Sutskever, I. 2019.
\newblock Language Models are Unsupervised Multitask Learners.
\newblock \emph{OpenAI Blog}.

\bibitem[{Rudman and Eickhoff(2024)}]{rudman2024stableanisotropicregularization}
Rudman, W.; and Eickhoff, C. 2024.
\newblock Stable Anisotropic Regularization.
\newblock arXiv:2305.19358.

\bibitem[{Shen et~al.(2025)Shen, Yan, Zhang, Hu, Du, and He}]{shen2025codicompressingchainofthoughtcontinuous}
Shen, Z.; Yan, H.; Zhang, L.; Hu, Z.; Du, Y.; and He, Y. 2025.
\newblock CODI: Compressing Chain-of-Thought into Continuous Space via Self-Distillation.
\newblock arXiv:2502.21074.

\bibitem[{Stiennon et~al.(2022)Stiennon, Ouyang, Wu, Ziegler, Lowe, Voss, Radford, Amodei, and Christiano}]{stiennon2022learningsummarizehumanfeedback}
Stiennon, N.; Ouyang, L.; Wu, J.; Ziegler, D.~M.; Lowe, R.; Voss, C.; Radford, A.; Amodei, D.; and Christiano, P. 2022.
\newblock Learning to summarize from human feedback.
\newblock arXiv:2009.01325.

\bibitem[{Su et~al.(2025)Su, Zhu, Xu, Jiao, Tian, and Zheng}]{su2025tokenassortedmixinglatent}
Su, D.; Zhu, H.; Xu, Y.; Jiao, J.; Tian, Y.; and Zheng, Q. 2025.
\newblock Token Assorted: Mixing Latent and Text Tokens for Improved Language Model Reasoning.
\newblock arXiv:2502.03275.

\bibitem[{Sui et~al.(2025)Sui, Chuang, Wang, Zhang, Zhang, Yuan, Liu, Wen, Zhong, Chen, and Hu}]{sui2025stopoverthinkingsurveyefficient}
Sui, Y.; Chuang, Y.-N.; Wang, G.; Zhang, J.; Zhang, T.; Yuan, J.; Liu, H.; Wen, A.; Zhong, S.; Chen, H.; and Hu, X. 2025.
\newblock Stop Overthinking: A Survey on Efficient Reasoning for Large Language Models.
\newblock arXiv:2503.16419.

\bibitem[{Uesato et~al.(2022)Uesato, Kushman, Kumar, Song, Siegel, Wang, Creswell, Irving, and Higgins}]{uesato2022solvingmathwordproblems}
Uesato, J.; Kushman, N.; Kumar, R.; Song, F.; Siegel, N.; Wang, L.; Creswell, A.; Irving, G.; and Higgins, I. 2022.
\newblock Solving math word problems with process- and outcome-based feedback.
\newblock arXiv:2211.14275.

\bibitem[{Wang et~al.(2024)Wang, Li, Shao, Xu, Dai, Li, Chen, Wu, and Sui}]{wang2024mathshepherdverifyreinforcellms}
Wang, P.; Li, L.; Shao, Z.; Xu, R.~X.; Dai, D.; Li, Y.; Chen, D.; Wu, Y.; and Sui, Z. 2024.
\newblock Math-Shepherd: Verify and Reinforce LLMs Step-by-step without Human Annotations.
\newblock arXiv:2312.08935.

\bibitem[{Wang et~al.(2023)Wang, Wei, Schuurmans, Le, Chi, Narang, Chowdhery, and Zhou}]{wang2023selfconsistencyimproveschainthought}
Wang, X.; Wei, J.; Schuurmans, D.; Le, Q.; Chi, E.; Narang, S.; Chowdhery, A.; and Zhou, D. 2023.
\newblock Self-Consistency Improves Chain of Thought Reasoning in Language Models.
\newblock arXiv:2203.11171.

\bibitem[{Wei et~al.(2023)Wei, Wang, Schuurmans, Bosma, Ichter, Xia, Chi, Le, and Zhou}]{wei2023chainofthoughtpromptingelicitsreasoning}
Wei, J.; Wang, X.; Schuurmans, D.; Bosma, M.; Ichter, B.; Xia, F.; Chi, E.; Le, Q.; and Zhou, D. 2023.
\newblock Chain-of-Thought Prompting Elicits Reasoning in Large Language Models.
\newblock arXiv:2201.11903.

\bibitem[{Yao et~al.(2023)Yao, Yu, Zhao, Shafran, Griffiths, Cao, and Narasimhan}]{yao2023treethoughtsdeliberateproblem}
Yao, S.; Yu, D.; Zhao, J.; Shafran, I.; Griffiths, T.~L.; Cao, Y.; and Narasimhan, K. 2023.
\newblock Tree of Thoughts: Deliberate Problem Solving with Large Language Models.
\newblock arXiv:2305.10601.

\bibitem[{Zhang et~al.(2025)Zhang, Zheng, Wu, Zhang, Lin, Yu, Liu, Zhou, and Lin}]{zhang2025lessonsdevelopingprocessreward}
Zhang, Z.; Zheng, C.; Wu, Y.; Zhang, B.; Lin, R.; Yu, B.; Liu, D.; Zhou, J.; and Lin, J. 2025.
\newblock The Lessons of Developing Process Reward Models in Mathematical Reasoning.
\newblock arXiv:2501.07301.

\bibitem[{Zhou et~al.(2023)Zhou, Schärli, Hou, Wei, Scales, Wang, Schuurmans, Cui, Bousquet, Le, and Chi}]{zhou2023leasttomostpromptingenablescomplex}
Zhou, D.; Schärli, N.; Hou, L.; Wei, J.; Scales, N.; Wang, X.; Schuurmans, D.; Cui, C.; Bousquet, O.; Le, Q.; and Chi, E. 2023.
\newblock Least-to-Most Prompting Enables Complex Reasoning in Large Language Models.
\newblock arXiv:2205.10625.

\bibitem[{Zhu et~al.(2025)Zhu, Peng, Cheng, Qu, Huang, Zhu, Wang, Xue, Zhang, Shan, Cai, Kergan, Kembay, Smith, Lin, Nguyen, Pan, Chou, Cai, Wu, Zhao, Liu, Yang, Zhou, Zheng, Li, Zhou, Li, Zhang, Liu, Zhang, Huang, and Eshraghian}]{zhu2025surveylatentreasoning}
Zhu, R.-J.; Peng, T.; Cheng, T.; Qu, X.; Huang, J.; Zhu, D.; Wang, H.; Xue, K.; Zhang, X.; Shan, Y.; Cai, T.; Kergan, T.; Kembay, A.; Smith, A.; Lin, C.; Nguyen, B.; Pan, Y.; Chou, Y.; Cai, Z.; Wu, Z.; Zhao, Y.; Liu, T.; Yang, J.; Zhou, W.; Zheng, C.; Li, C.; Zhou, Y.; Li, Z.; Zhang, Z.; Liu, J.; Zhang, G.; Huang, W.; and Eshraghian, J. 2025.
\newblock A Survey on Latent Reasoning.
\newblock arXiv:2507.06203.

\end{thebibliography}

\end{document}